# Improving Multilingual Neural Machine Translation System for Indic Languages

SUDHANSU BALA DAS, National Institute of Technology (NIT), Rourkela, India
ATHARV BIRADAR, Pune Institute of Computer Technology (PICT), Pune, India
TAPAS KUMAR MISHRA, National Institute of Technology (NIT), Rourkela, India
BIDYUT KUMAR PATRA, Indian Institute of Technology (IIT), Varanasi, India

Machine Translation System (MTS) serves as an effective tool for communication by translating text or speech from one language to another language. Recently, neural machine translation (NMT) has emerged to be popular for its performance and cost-effectiveness. However, due to the large quantity of data needed to learn useful mappings across languages, NMT systems are restricted in translating low-resource languages. The need of an efficient translation system becomes obvious in a large multilingual environment like India, where English and a set of Indian Languages (ILs) are officially used. In contrast with English, ILs are still entreated as low-resource languages due to unavailability of corpora. In order to address such asymmetric nature, multilingual neural machine translation (MNMT) system evolves as an ideal approach in this direction. MNMT, which converts many languages using a single model, is extremely useful in terms of training process, lowering online maintenance costs, and improving low-resource translation. In this paper, we propose a MNMT system to address the issues related to low-resource language translation. Our model comprises of two MNMT systems i.e. for English-Indic (one-to-many) and the other for Indic-English (many-to-one) with a shared encoder-decoder containing 15 language pairs (30 translation directions). Since most of IL pairs have scanty amount of parallel corpora, not sufficient for training any machine translation model. We explore various augmentation strategies to improve overall translation quality through the proposed model. A state-of-the-art transformer architecture is used to realize the proposed model. Trials over a good amount of data reveal its superiority over the conventional models. In addition, the paper addresses the use of language relationships (in terms of dialect, script, etc.), particularly about the role of high-resource languages of the same family in boosting the performance of low-resource languages. Moreover, the experimental results also show the advantage of backtranslation and domain adaptation for ILs to enhance the translation quality of both source and target languages. Using all these key approaches, our proposed model emerges to be more efficient than the baseline model in terms of evaluation metrics i.e BLEU (BiLingual Evaluation Understudy) score for a set of ILs.

## 1 INTRODUCTION

Numerous languages persists around the globe. Interestingly, every language has its own lexical interpretation with a vast vocabulary (words and phrases) and rules (grammar) that differs from other. Except some special cases, it is not possible for an individual to be familiar with more than one language. For this reason, it becomes difficult to understand an unfamiliar language without proper translation/interpretation, which is generally done by a third-party i.e. Human. Due to factors like affordability, dependency, human-error, time and others in case of a human interpreter, machine translation has been the preferred choice [22]. India is home to numerous ancient and morphologically rich languages used distinctly over various regions. Apart from having a set of dialects and accents, every regional language has its own identity and use. On the contrary, English is the most common language across the country as a medium of sharing information among citizens not only for administrative works but also for exchanging sentiments, emotions, ideas and actions over global media (social platform). In the vast sector of Information Technology, English is preferred to other natural languages since the information

Authors' addresses: Sudhansu Bala Das, National Institute of Technology (NIT), Rourkela, Odisha, India, 769008, 520cs6006@nitrkl.ac.in; Atharv Biradar, Pune Institute of Computer Technology (PICT), Pune, India; Tapas Kumar Mishra, National Institute of Technology (NIT), Rourkela, Odisha, India, mishrat@nitrkl.ac.in; Bidyut Kumar Patra, Indian Institute of Technology (IIT), Varanasi, Uttar Pradesh, India, bidyut.cse@iitbhu.ac.in.

provided in the English language by the standard ASCII symbols is relatively simple for computers to process [60]. However, those who are less conversant with English find it difficult to cope with and often need proper translation/interpretation for clarity at every step. In order to bridge the gap, a machine-based translation is a perfect approach that can achieve all these tasks with little human involvement.

Machine Translation (MT) generally refers to an independent process of translating, primarily through a computer application, from one language as source/input to another language(s) as target(s)/output. With the advent of natural language processing (NLP), a subset of artificial intelligence (AI), computers detect and sense the intent of input language, and translate accurately as per the format of output language. Such process is very effective in terms of time (speed), volume and cost [3] . In consideration with Indian environment, as discussed earlier, the development of a quality machine translation system (MTS) for the ILs is in huge demand. Still, it remains as a challenging task, since numbers of ILs are individually low in terms of resources resulting in an adverse impact on their translation quality. Therefore, translation patterns, both from English to IL and from IL to English, face problems in morphological as well as structural analysis. Even, the variation in word-orders and the dissimilarity in the sentence size (of the source and target) create an issue in the word alignment and reduce the translation quality. However, recent research propose that proper usage of the parallel and monolingual corpus increases the translation quality of low-resource language. Neural Machine Translation (NMT) system[5], [69] have gained popularity and have shown better results than the Statistical Machine Translation (SMT)[31] System. Although a traditional NMT model can successfully translate between single language pair (for example, Hindi and English), yet training a discrete model for every language pair is impractical due to a large number of languages spoken worldwide. This problem is resolved with the recent development of multilingual neural machine translation (MNMT) model through various approaches. One such approach is transfer-learning where a model is first trained on a high-resource language pair, and then parameter values are transcribed from that model and further fine-tuned on low resource data [48]. This technique helps to increase the amount of corpus as well as the translation quality for low resource languages. Hence, the main objective of MNMT system is to utilize all the corpora of different language pairs and make one model which can give a qualitative translation [9]. Recent research-works on MNMT evoke promising methods for improving quality with low-resource languages trained in a multilingual setting. It also works well with translation between languages that do not have any parallel corpus at the time of training i.e., zero-shot translation [9]. Both the source and target languages are included in the training and the MNMT system can translate between unknown pairs automatically without any manual intervention in this mode of translation. In case of low-resource languages, lexical and orthographic relationships among languages may be used to make the quality of translation better [35]. The links made between words are known as lexical relationships whereas orthographic relationship addresses punctuation rules and the link between spoken and written language.

In the domain of ILs being restricted with low-resource, the use of MNMT, with some associated strategies, is preferred to generate an ideal model for obtaining a qualitative translation system. Applying a string of procedures over a transformer [69] based MNMT model makes it possible which is thoroughly described in this paper. Whole parallel, as well as monolingual corpus of Samantar dataset [55] is utilised to build a MNMT system for fifteen language directions. The translation outputs are also reviewed using automatic evaluation metrics i.e BLEU score [52].

### 1.1 Motivation

Machine translation (MT) systems have significantly improved in recent years, but only with the condition of abundant parallel data availability [22]. However, the MT issue for many low-resource languages remains in situ without remarkable change. Out of nearly 7100 spoken languages worldwide, just a small number languages have access to automatic translation tools [39]. Such trend hinders a seamless exchange of knowledge in global context which in turn becomes a major obstacle for the advancement of humanity as a whole. As a matter of fact, many Indians find it difficult to cope with such trend since ILs, being low-resource, are hardly used on global platforms. Therefore, a qualitative MTS for ILs is in huge demand which motivates us for this work. Numerous strategies, including unsupervised, semi-supervised, and multilingual learning have been developed to address the low-resource issue [44]. Monolingual data is used for unsupervised and semi-supervised methods, whereas the multilingual MT gains from the knowledge transfer caused by the fusion of various languages. It has been discovered that knowledge transfer from high-resource languages benefits low-resource languages [54]. Eventually, it motivates us to explore an ideal MTS for ILs, with optimum quality, employing state-of-the-art neural architecture through a novel approach. For this, MNMT with transformer model is utilised with extensive training for ILs over 30-directions in combination with English (EN). Various processes like noise reduction, language relatedness, back-translation and domain adaptation are also incorporated towards a qualitative output.

### 1.2 Our Contribution

Our novel approach towards realisation of a qualitative MT system for ILs to English (and back) comprises of various evaluation processes with satisfactory test-results. Moreover, in the course of this research work, following major contributions have been achieved:

(1) As a first attempt ever, to the best of our knowledge, this novel work leads to explore the MT of fifteen EN-IL and IL-EN pairs (both directions), including both the Indo-Aryan and Dravidian language groups, in a multilingual environment using cutting-edge transformer architecture, and to assess their effectiveness.
(2) Apart from trials over the linguistic features commonly occur in the ILs, different data filtration techniques are explored to clean the data for betterment of translation quality.
(3) We check language relationship approach in MT System which plays a major role in Indic languages. It is noticed that language mix-up techniques (based on IL similarity i.e., script, dialects etc.) help the low-resource languages to achieve better quality translation.
(4) We also examine the effectiveness of back-translation and domain adaptation technique in achieving better results in translating both low-resource and high-resource ILs.
(5) This paper establishes the performance superiority of our model through the results in comparison with fine-tuned OPUS-MT (pretrained) model for majority of IL pairs using PMI corpus.

This paper is organised as follows. In section 2, we concisely describe some prominent work on Multilingual Machine Translation (MNMT) systems. Section 3 talks about the approaches used; and Section 4 discusses the architecture of the model, details about the corpus, pre-processing steps, characteristics of ILs and the procedure of training the model. Section 5 explains about the model overview and all subsequent training approaches, whereas in section 6, the performance of finetuning OPUS model with our model is compared. Results are shown in Section 7, followed by the conclusion in Section 8.

## 2 RELATED WORK

The first machine translation (MT) study was undertaken in the year 1950s, and since then, a significant amount of work has been reported on MT [25]. The MT system initially determines the translation of a text in the source language by matching words and its meanings in the source language to those in the target language with set of rules. Researchers utilize a variety of methodologies, including rule-based [17], corpus-based [61] [10], and hybrid-based techniques [56]. There are advantages and disadvantages to each strategy. Due to their inability to capture the diverse sentence structures found in the language, these approaches could not produce satisfactory results. This lengthy translation process also calls for individuals who are fluent in both languages. In order to address the shortcomings of rule-based systems, corpus-based translation approaches such as statistical machine translation (SMT) [31] and NMT [5] [7] [64] have been developed. In SMT[63], the preprocessed data are examined with statistical metrics to produce the desired outcome in language processing tasks. This method searches for statistical relationships in pre-process data (such as probability and distance metric, etc.). When translating a document, the probability distribution function **P(m/n)** is used as the basis for translation. The probability of converting a sentence **n** from the source language **N** (for example, English) to the sentence **m** target language **M**(example:Telugu) is represented by the **P(m/n)**. Due to the lack of high-quality parallel corpus, a lot of research has been done over machine translation from English to IL, primarily relying on rule-based techniques. Significant efforts have been done on using statistical and hybrid approaches to translate text from English to IL(s) despite the lack of appropriate parallel corpora. In SMT tasks, yet the outcomes are not noticeable. NMT, on the other hand, is a new technique that has significantly improved translation outcomes but can be used for one-to-one translation languages. Most NMT systems are supervised deep learning systems, which are extremely data-hungry. Despite years of research, only a small number of languages spoken worldwide has access to high-quality annotated MT resources [51]. So, one of the major issues that arises from MT's multilingualism and linguistic diversity is data scarcity. As per Naveen et al. [4] and G Lample et al. [36], the types of languages utilised to train the model have a significant impact on the effectiveness of the supervised MNMT models. MNMT systems have the capability to train a single model for multiple language pairs at a time. The first MNMT [14] is based on a Multiple-task learning framework used for translating in one-to-many language direction where for target languages, different language-dependent decoders and attention mechanisms were applied. Their approach to one-to-many language showed better results than individual language translation. Then, many-to-many language model for multilingual machine translation system was proposed by Firat et al. [16]. Their model is based on a shared attention mechanism with many numbers of encoders and decoders. In case of enhancing the quality of low resource language, Aharoni et al. [2] added transfer interference trade-off and they found that it is more efficient in many-to-one(English) direction. To boost the performance of the MNMT models, there are different ways that work significantly with the training method as well as with the model architecture of Wang et al. [71], Aharoni et al. [2], Lin et al [40]. Approaches on augmentation of the corpus (both parallel as well as monolingual) like back-translation, transliteration, etc. for low resource language enhance the quality of translation in MNMT model [50]. A brief summary of all the techniques, architecture selection, and other parameters relevant to MNMT as well as issues related to it was described by Dabre et al [9]. In recent scenarios [15], pre-trained language models an multiple languages have proved favourable for the MNMT system. In case of low-resource languages like IL, many researchers [1], [62] have trained and tested MNMT models on IL corpora from online website in different domains. An extensive MNMT model that can work with 102 languages was proposed by Aharoni et al [2] which focuses on training models on the corpora of many language pairs, with English as a source or target language. Even with incredible progress of MNMT, it is sensitive to the noise in the corpus[6]. Hence, various filtering techniques have gained popularity and have become crucial to clean the corpus and to remove the unwanted contents, symbols, etc. Liu et al[41], Pinnis et al[53] and Li et al [38] have experimented different filtering techniques with parallel corpora and then trained multilingual models to achieve progressive results. mBART [42] is the first pre-trained multilingual model based on sequence-to-sequence architecture. In mBart model, filtered corpus of different languages (from noises) are used to train the model for multiple languages, achieving outstanding results in terms of BLEU score. This shows the importance of noise removal for a substantial enhanced performance of both supervised and unsupervised MT, applicable of both sentences level and document level. So, lots of techniques, approaches and methods have been used by researchers to achieve translation quality.

## 2.1 Background

This section gives background information with particular emphasis on traditional bilingual NMT and an MNMT systems.

*2.1.1 Neural Machine Translation(NMT) System.* A radical improvement over earlier MT techniques is NMT. In addition to embracing the probabilistic framework, NMT offers a data-driven approach to MT [9]. Provided a parallel dataset $C$ the NMT reduces the translation task into the probability distribution $p$ of a target language language $b$ given source language $a$ :

$$p(b \mid a; \omega) = \prod_{i=1}^{y} p\left(b_i \mid b_{(i-1,\ldots,1)}, a; \omega\right) \tag{1}$$

where $a$= $a_1$,.....,$a_x$ is an input source language sentence of $x$ words whereas the translated sentence i.e the target language of $y$ words is $b$, $b$ =$b_1$,.....,$b_y$, $\omega$ is the parameter to be learned, $b_i$ is the presently produced word and $b_{(i-1,\ldots.1)}$ are the previously created word. The log-likelihood $\mathcal{S}$ with respect to the parameter set $\omega$ is maximized during training of model of Eq (1) by the Eq (2):

$$\mathcal{S}(\omega) = \sum_{(a,b)\epsilon C} \log p(b \mid a; \omega) \tag{2}$$

In contrast to the early studies on NMT which focused on developing translation systems between bilinguals, nowdays researchers found that the NMT framework can naturally include numerous languages. This approach has demonstrated cutting-edge performance for different language pairs. As a result, research on Multilingual MT systems has significantly increased [9].

*2.1.2 Multilingual Neural Machine Translation(MNMT) System.* MNMT systems are capable of translating between multiple language pairs [9]. It significantly increases the translation quality of low resource languages. The low resource languages learn additional information by training on high resource languages. MNMT training can be done in a variety of ways:

- Transfer Based [73]: Translation knowledge transfer learning uses learned attributes from well resource language pairs to train related language pairs with less resources .
- Pivot Based [8][45] [68]: When direct parallel data between source and target language are unavailable, pivot based system uses a pivot language $p$ to link the translation between the two.
- Multiway Based [9]: Using parallel corpora for several language pairs, the aim of multiway translation is to build a single NMT system for many-to-one, one-to-many, or many-to-many translation.

MNMT's main objective is to create a model that facilitates translation between multiple language pairs i.e any source-and target-language pairs i.e $N_{(sourc)_m}$,$N_{(targt)_m}$ respectively. $M$ is the total number of acceptable language pairs. Eq (2) is used in Eq (3) to compute loss of individual language pairings $S^{N_{(sourc)m},N_{(targt)m}}$ where m is the currently used language pair. The goal of multiway NMT training is to maximize the log-likelihood $S_\omega$ (where $\omega$ is the parameter to be learned) of all training data taken collectively for all language pairs (different weights may be assigned to the likelihoods of different language pairs) as shown in Eq (3).

$$\mathcal{S}_\omega = \frac{1}{M} \sum_{m=1}^{M} \mathcal{S}^{N_{(sourc)m},N_{(targt)m}}(\omega) \tag{3}$$

Moreover, MNMT may be classified into three types depending on the alignment of the source and target languages namely; **a) Many to Many [16]**: Translation between a wide variety of source and target languages is feasible in this category. **b) One to many [14]**: A single source language is translated using the MNMT approach into numerous target languages. **c) Many to One [37]**: The model is trained to translate from various source languages into a single target language in this context. The next section describes about the approaches which we have used to build our MT system for 15 EN-IL and IL-EN pairs i.e., 30 directions.

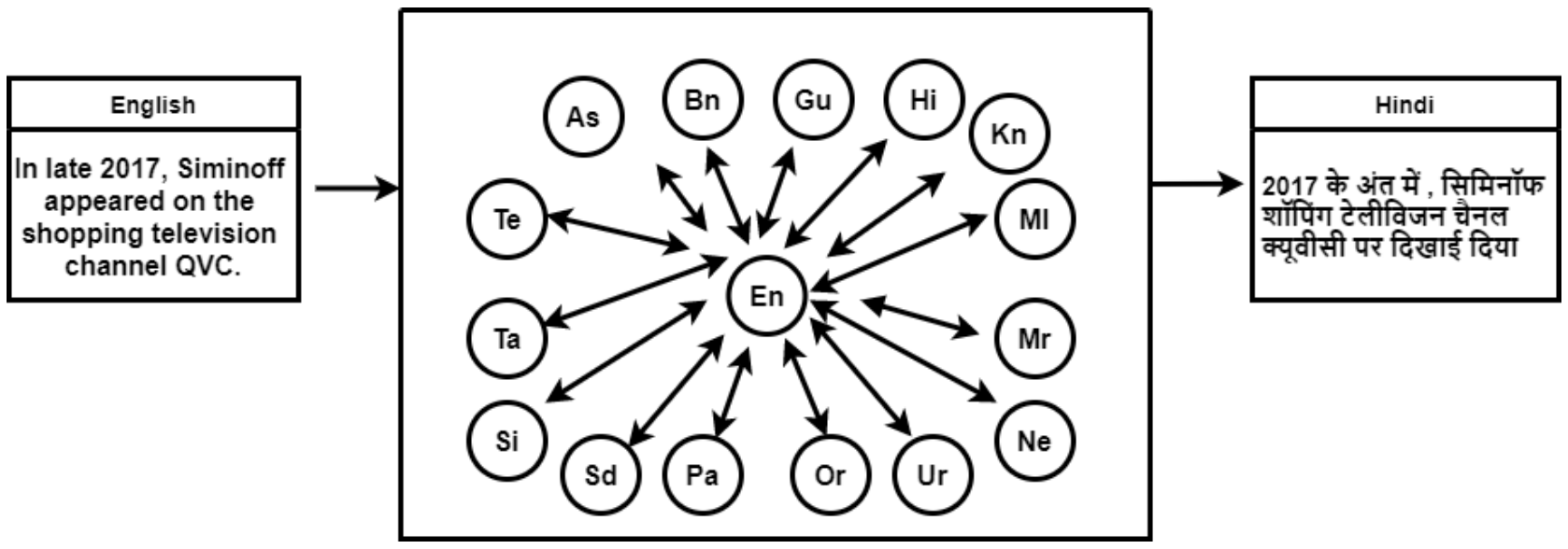

Fig. 1. One to Many and Many to One MNMT System

## 3 APPROACH USED

Previous studies on MNMT focuses on building several encoders and decoders for every source/target language. As per cutting-edge research by Johnson et al [26]. MNMT model does not alter the original NMT model, and instead employs a single NMT model to translate between multiple languages. It also affirms that mixing low-resource languages with high-resource languages significantly improve translation effectiveness on low-resource bitexts. Hence, we use Johnson et al [26] approach of adding an artificial token at the start of the sentence to display the target language that the model wanted to achieve. Fig. 1. describes one-to-many and many-to-one translation with one MNMT model. In case of Indic-to-English language pair, we use a single shared encoder and one decoder on the target language (English), which we need to translate for all source languages. For English-to-Indic translation task, we use a single encoder for the source (English) and a shared decoder for target languages. In this approach, we have added token that indicate a target language (which we need as the output of translation) at the beginning of the source language sentence.
For example : let's say, English to Hindi sentence pair :**Where are you?** can be written in hindi as : आप कहाँ है ?
The above instance shows a simple example of NMT.
Similarly, for multilingual NMT, it will be converted as: **<2hi>Where are you?** and in hindi it can be written as <hi> आप कहाँ है ? . One thing we can notice here is the token i.e <hi>, which indicates the target language whereas the source language is not mentioned anywhere and model learns itself during training. Once introducing the token to the input data is done, we train the model using all multilingual data containing multiple languages at once, after accounting for the relative ratio of language data available.

## 4 PROCESSING OF INDIC LANGUAGES FOR TRANSLATION

This section illustrates the details architecture of the system, corpus/dataset, data preprocessing and filtering steps and model training details.

### 4.1 Architecture

Proposed MT system has been realized with transformer architecture as described by Vaswani et al [69], where at each step it assigns self-attention mechanism which passes the information among the encoders and decoders efficiently as shown in Fig. 2. Indeed, our system is designed in consideration with following basic advantages of transformer architecture over other NMT architectures in a low-resource MNMT scenario [72]. Transformer models are highly parallelizable, which makes them incredibly compute-optimal and enables to train extremely big models [69]. In order to establish dependency between input and output, transformer architecture relies on self-attention instead of recurrence and convolution. Such mechanism provides transformer architecture direct access to data from every step unlike sequential models [72]. Apart from having a traditional six-encoder-decoder architecture, the transformer primarily relies on attention layers to translate source sentences into target sentences. It employs stacked encoders, each of which has two sublayers: a self-attention layer and a feed-forward neural network (FFNN), to learn the representations of source sentences. The importance of each token in the self-attention layer is learned by analyzing the other appropriate tokens in the sentence. The decoding side allows the decoder to pay focus on particular features in the source sentence encoding during translation. Since a transformer model can handle all the data concurrently, both past and future elements are handled at the same time. This leads to less processing time and more effective training (prime requirement for a low-resource language). For our novel method of realising a qualitative MTS towards ILs-English (in both directions) using transformer architecture, the complete assignment is done employing 6 encoder-decoder layers with 8 attention heads, 512 embedding dimensions and 2048 layer FFNN.

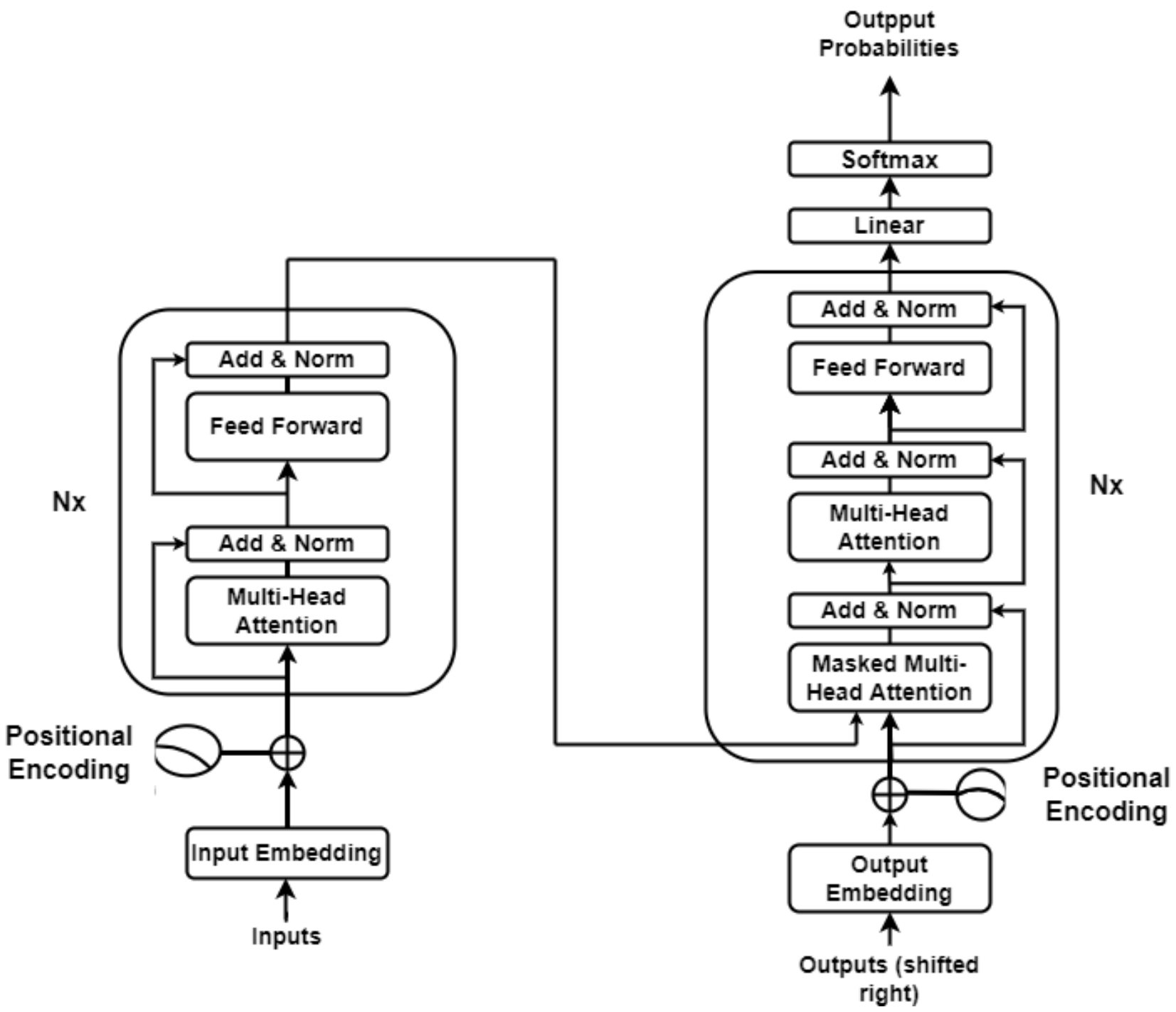

Fig. 2. Transformer Architecture [69]

## 4.2 Dataset

Samanantar Corpus [55] are used to train our Multilingual Machine Translation (MNMT) system. This corpus includes more than 40 million sentence pairs between English to indic languages wherefrom Assamese(as), Malayalam(ml), Bengali(bn), Marathi(mr), Gujarati(gu), Kannada(kn), Hindi(hn), Oriya(or), Punjabi(pa), Telugu(te) and Tamil(ta) are used. The parallel data included in corpus are collected from various sources such as: PMIndia, CVIT-PIB, UFAL EnTam, IITB 3.0,Uka Tarsadia, JW, NLPC, Wiki Titles, ALT, OpenSubtitles, Bibleuedin, MTEnglish2Odia, WikiMatrix , OdiEnCorp 2.0, TED [47]. For Nepali(ne), Sindhi(sd), Sinhala(si) and Urdu(ur) language, we use OPUS corpus [65]. For Nepali and Sinhala, Flores101 [18] and FLoRes [21] dataset are used for testing purpose. For testing purpose, Flores101 [18] and FLoRes [21] dataset (only for Nepali and Sinhala) are used. No other corpus or any other resources related to linguistic were used for our experiments. Table 1 gives full detail about the both monolingual [28] and parallel [55] corpus where NA represents Not Available.

Table 1. Parallel and Monolingual corpus statistics

| English to Indic | Parallel Corpus(Sentences) | Monolingual Corpus (Sentences ) |
|---|---|---|
| Bengali(bn) | 8.52M | 39.9M |
| Gujarati(gu) | 3.05M | 41.1M |
| Hindi(hn) | 8.56M | 63.1M |
| Kannada(kn) | 4.07M | 53.3M |
| Malayalam(ml) | 5.85M | 50.2M |
| Marathi(mr) | 3.32M | 34.0M |
| Oriya(or) | 1.00M | 6.94M |
| Punjabi(pa) | 2.42M | 29.2M |
| Tamil(ta) | 5.16M | 31.5M |
| Telugu(te) | 4.82M | 47.9M |
| Assamese(as) | 1.4M | 1.39M |
| Urdu(ur) | 6.1M | NA |
| Nepali(ne) | 0.7M | NA |
| Sinhala(si) | 6.3M | NA |
| Sindhi(sd) | 1.7M | NA |

### 4.3 Linguistic Features of Indian Languages [10], [11]

IL exhibit a rich morphology [70]. Due to this, IL present difficulties for MT and NLP tasks. The issues of structural and morphological variations exist when translating from English to ILs. The word order is a key structural distinction between English and ILs [32]. While most ILs primarily utilize subject-object-verb(SOV), English uses the subject-verb-object(SVO) sequence. Also, some ILs have unrestricted word order by nature. Therefore, when translating, a translator i.e human or machine—must use the transformation between VO and OV. Few of the linguistic features of ILs are:

- Duplication: This refers to a phenomenon where a word is used in repetition to convey a variety of speech acts, including intensity, plurality, emphasis etc. For example, we can write: ‘at village village (in English language)’. becomes गाँव गाँव में in hindi. English duplication can be ‘at villages’, and it is evident that the term ‘village’ cannot be repeated twice in the English sentence. Here, the reduplication of गाँव implies plurality i.e., villages. As reduplication happens to be a normal occurrence in ILs, translating the same into another IL is as simple as a replication of the original language. As a result, it is simpler to manage redundancy transfer across ILs than it is for languages other than those. For example, the same is translated in Odia ଗାଁ ଗାଁରେ . Therefore, a little extra effort is required in the translation process in order to properly manage such duplication for an actual representation while translating between En and IL.
- Untranslatability: Interestingly, some words/phrases in one language do not have the equivalent direct translation in other languages. For example, oxymorons like ‘Hauntingly beautiful’ translate in bengali ভুতুড়ে সুন্দর and ਹੰਕਾਰ ਨਾਲ ਸੁੰਦਰ (using google translator) which hardly carries any sense. One more example, टेढ़ी खीर in hindi translate in english as ‘crooked cake’ but it means ‘difficult work’. In addition, direct translation of sarcastic/hyperbolic expressions(those differ from one language to other) often makes the sentence lose its original meaning. For example, अच्छा used sarcastically in Hindi is translated to 'really' in English despite of its direct translation being 'OK/well'.

- Ambiguity: In many instances, a particular word in a language has different meanings based on the context/manner it is used in a sentence. Direct translation (to another language) of such words, without much analysis, often conveys a different meaning with slight or complete variation. For example the word 'North' means उत्तर in Hindi, but the same word may represent 'answer' or 'direction' in other instances- completely different in meaning.
- Context: Based on situation and context, a word may mean differently depending on its use in the sentence. For example, the word 'spot' means 'position' in the sentence "you've put me in a tough spot", whereas the same word refers to 'see/find' in the sentence "spot the bird".
- Synchronous Digraphia: Refers to a situation where a single language possesses more than one script, making translation process quite difficult, particularly for training the model. In such case, two tokenizers are required to tokenize the language, or the scripts need to be transliterated before passing to the model. Example of this in ILs is Konkani where two scripts i.e., Devanagari and Kannada, are used.
- Gender: Various languages have different notions of gender and distinctive means they are interpreted in. In some languages (Odia/English/Bengali), verbs possess no genders unlike languages like Hindi and Urdu. In languages like Marathi, Konkani etc., there exists a third (neuter) gender; and in some others, genders also exist to define animate and inanimate objects (e.g., Tamil).
- Compound Words and Agglutination: These are the grammatical structures where two or more words are combined to make a single word which may have completely different meaning. For instance, पीत in Hindi means 'yellow', अंबर in Hindi means 'cloth', but पीताम्बर refers to 'Lord Krishna'. Even in English, we have similar examples like ice-cream, doorknob, suitcase. Agglutination is similar to compound words, but it is made of suffixes (meaningless parts) instead of words.

### 4.4 Transliteration for Similar Languages

Transliteration is an approach used for transforming a word from an original text to a target language that frequently uses MT techniques. The purpose of transliteration is to retain as much of the original pronunciation of the source word as possible while adhering to the phonological structures of the target language. For example, the sentence in Hindi (Devanagari) हम किताबें पढ़ते हैं is transliterated as **"hum kitaaben padhte hain"** in English. A transliteration system **T** accepts a source word **S** and returns a ranked list **R** with ($M_i$, $Ki$) tuples as its elements. $M_i$ is the **i**th transliteration of the source word S obtained with the $ith$ highest likelihood $Ki$ in each tuple. There are numerous scripts used for Indian languages. Transliteration across scripts of related languages, as discussed by Haddow et al [23], Goyal et al [20], may enhance the quality of multilingual models. This approach is used in similar languages determined based on the similarity in the languages [38] such as their dialects, script, word order etc. We have used related-language transliteration technique for the languages that fall into a similar group. For transliteration, we utilised the Indic NLP Library [34]. For instance, Odia, Bengali, and Assamese are the languages that have similarities. So, Bengali training data i.e., high resource language is transliterated to Odia and Assamese i.e., low resource; and then gets enhanced using training data of Oriya and Assamese Language. The training data for low resource languages has been supplemented with relatively high resource-related language training data that have been transliterated into the low resource language. Similarly, Gujarati and Punjabi which are low resource languages are similar to Hindi. Therefore, Hindi training data is transliterated to Gujarati and Punjabi and then added to training data of Gujarati and Punjabi.

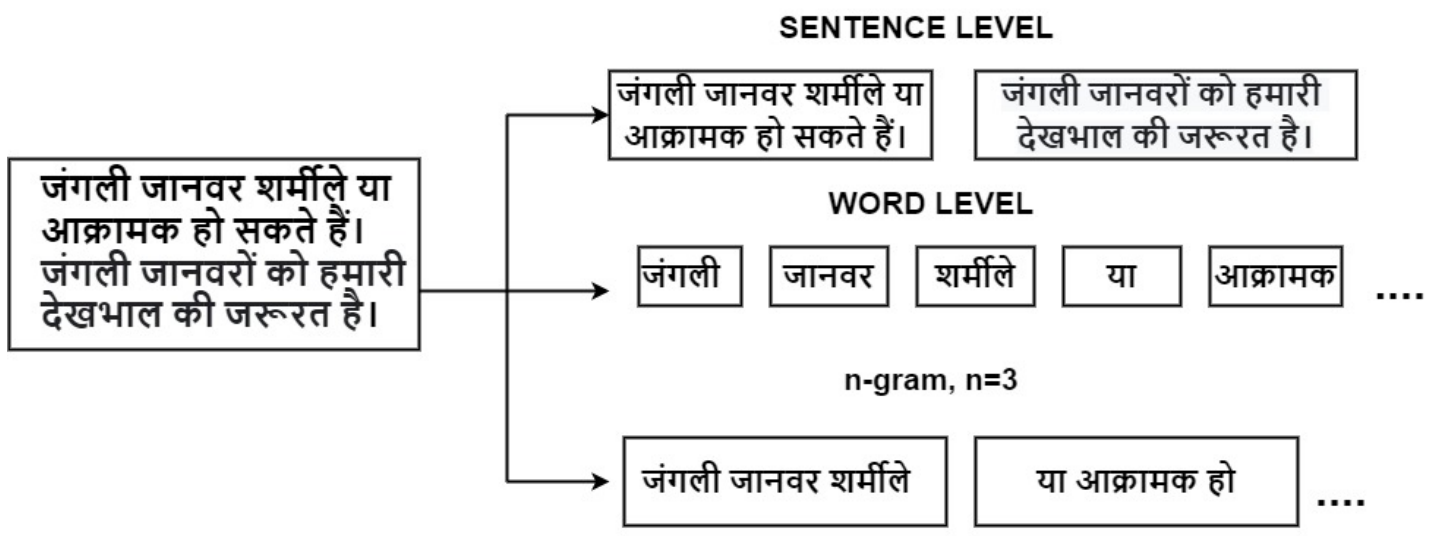

Fig. 3. Example of Tokenization using Hindi Language

## 4.5 Data Preprocessing

All our experiments are done using Byte Pair Encoding (BPE) which is an efficient technique for data division i.e splitting up the words into sub-words [59]. BPE works very well for Indian languages having morphological richness features. After normalization and pre-tokenization are complete, BPE training begins by computing the distinct set of words used in the corpus. The vocabulary is then built by utilizing all of the symbols used to represent those words. BPE has the advantage of making UNKs obsolete. In MT tasks, UNK symbol denote the words which are not present in vocabulary. Before learning the BPE codes, the data from all 15 Indic languages is merged for training the one-to-many and many-to-one models. The BPE codes of the one-to-many and many-to-one are learned using 48000 and 6400 merge operations, respectively.

## 4.6 Tokenization

The initial step in any machine translation applications involves tokenizing the raw text that involves tokenizing the given text into lexical units, which are the most fundamental components [13]. Each lexical unit is designated as a token after tokenization. Depending on the type of issue, tokenization may occur at the phrase or word level. Three different types of tokenization are: a) tokenization at the sentence level, b) word level tokenization, and c) Subword level tokenization as shown in Fig. 3. Tokenization at sentence level addresses issues like ambiguity, word sequences and the detection of sentence endings; whereas in the word level tokenization, words serve as lexical units. The entire document is tokenized to a set of words as the applications such as language processing and text processing frequently use the word level of tokenization. The n-gram tokenization is an n-words token, where 'n' is the lexical units of total number of words. If 'n' is equal to 1, 'n' is a unigram, and 'n' is equal to 2, then 'n' is a bigram, and 'n' is equal to 3 then 'n' is a trigram. Instead of using word or character level tokenization, we use sentence piece tokenizer to take the advantage of the morphological richness attribute of ILs. SentencePiece enhances direct training using raw sentences to include sub-word units and a unigram language model. We create a fully end-to-end system using SentencePiece that is independent of language-specific processing (both pre and post). SentencePiece tokenizer [33] combined with 48K vocabulary of 15 target IL along with English has been used for En-XX Multilingual Translation wherein character coverage of 1.0 has been utilized. For XX-En, sentencepiece tokenizer [33] has been employed using a combined vocabulary of 64K words from the 15 source Indic languages with 1.0 character coverage.

## 4.7 Data Filtering

Currently there are an increasing number of parallel corpora accessible for MT training [66]. However, we can never be certain about data quality when obtaining corpora from different sources, which is crucial for an MT system's effectiveness [29]. A short scan of Samantar corpus used for translation task (which is not noise-free), just like the majority of corpus used for MT tasks provides an overview of the noises and helps in the application of a set of heuristics to remove many of those noisy sentence pairings. Many methods/experiments have been done to enhance and filter Samanantar Corpus [55]. Few of the methods used are:

- Removing the sentence pairs where the source or the destination language is empty.
- Removing low likelihood sentence pairs (based on factors like sentence lengths, identified languages, etc.)
- Removing sentences containing characters from a certain language pair's Unicode range.

Table 2. Filtered Parallel corpora statistics

| EN to Indic | Filtered | After Filtration (sentences) |
|---|---|---|
| en-bn | 11.53% | 7537644 |
| en-gu | 6.87% | 2840465 |
| en-hi | 8.21% | 7857224 |
| en-kn | 6.92% | 37788356 |
| en-ml | 8.71% | 5340465 |
| en-mr | 6.44% | 3106192 |
| en-or | 3.12% | 968800 |
| en-pa | 3.86% | 2326588 |
| en-ta | 7.62% | 4766808 |
| en-te | 7.45% | 4460910 |
| en-as | 4.24% | 1340640 |
| en-ur | 9.46% | 5522940 |
| en-ne | 2.56% | 682080 |
| en-si | 9.72% | 5687640 |
| en-sd | 5.03% | 1614490 |

Table 2 describes about filtered parallel corpora statistics. After all Indian scripts get standardized, tokenized, and transliterated using the Indic NLP library [35], experiment is carried out using both methods: a) by using noise free corpus, and b) without removing the noisy corpus. It becomes clear that none of these methods used on the parallel corpus lead to a significant quality improvement. Hence, the filtered corpus is preferred for translation task using different approaches.

## 5 MODEL OVERVIEW

An efficient strategy to increase the number of training instances for machine translation (MT) is data augmentation which is becoming a standard procedure for MTs with limited resources. In the next subsection, different techniques used in our experiments, to check the quality of translation, have been described.

### 5.1 Model Primal Training

In order to train the system, we have constructed two distinct MNMT models: (a) one (English) to many (15 IL) model and, (b) many (15 IL) to one (English) model. The transformer model [69] is employed in our one-to-many approach, with a single shared encoder and decoder. The decoder employs a shared vocabulary of all the Indic languages whereas the encoder uses the English language's vocabulary. For many-to-one model, the transformer architecture is used with a single shared encoder and a single decoder. Here, the shared vocabulary of all the IL has been employed for the encoder, and English vocabulary is used for the decoder. A token specific is added to the input language for both of these MNMT models before the phrase. For the implementation of the multilingual system, the fairseq [49] library is preferred. Adam [30] optimizer is utilized for training with betas of (0.9,0.98). With 8000 warm-up updates and an initial learning rate of 5e-0.4, the inverse square root learning rate scheduler has been employed. The criterion used is label smoothed cross entropy with a label smoothing of 0.1, and the dropout probability value has been set to 0.6. For the multilingual models, we have adopted an update frequency of 15 and have applied the beam search algorithm during the decoding process, with a beam length of 20. The one-to-many model has been trained for 12 epochs, whereas the many-to-one model has received 13 epochs with the training time of nearly 124 hours. Based on the correctness of the validation set, all of our models have been trained with early stopping criteria. During testing, we have reassembled all the translated BPE segments which have been changed back to the original language scripts. Finally, BLEU(Bilingual Evaluation Understudy) is used to assess the precision of our translation models [52]. The results of the primal training compared with baseline can be found in Table 3 and in Fig. 4 and Fig. 5, where it is found that 5 new languages corpus are added with no existence of baseline result for these languages [12].

Table 3. Primal Model Training Results with evaluation metrics as BLEU score

| | **Baseline** | **Proposed Primal Model** | | **Baseline** | **Proposed Primal Model** |
|---|---|---|---|---|---|
| **EN->Indic** | | | **Indic->EN** | | |
| en-bn | 15.97 | **27.40** | bn-en | 31.87 | 30.2 |
| en-gu | 27.80 | **29.90** | gu-en | 43.98 | 35.3 |
| en-hi | 38.65 | **39.9** | hi-en | 46.93 | 44.2 |
| en-kn | 21.30 | **26.10** | kn-en | 40.34 | 37.0 |
| en-ml | 15.49 | **20.90** | ml-en | 38.38 | 37.4 |
| en-mr | 20.42 | **21.90** | mr-en | 36.64 | **36.7** |
| en-or | 20.15 | **21.20** | or-en | 37.06 | 33.8 |
| en-pa | 33.43 | **35.90** | pa-en | 46.4 | 40.6 |
| en-ta | 14.43 | **18.90** | ta-en | 36.13 | **41.9** |
| en-te | 16.85 | **28.0** | te-en | 39.8 | **43.3** |
| en-as | - | **16.6** | as-en | - | **27.3** |
| en-si | - | **15.3** | si-en | - | **15.7** |
| en-ne | - | **6.6** | ne-en | - | **14** |
| en-ur | - | **28.70** | ur-en | - | **26.9** |
| en-sd | - | **15.50** | sd-en | - | **18.5** |

Table 4. Similar Languages

| Group | Languages Belong |
|---|---|
| A | Hindi, Urdu, Punjabi, Gujarati, Marathi, Oriya, Bengali, Sindhi |
| B | Telugu, Tamil, Kannada, Malayalam |

## 5.2 Subsequent Training Approaches

As the Table 5 shows the comparison of Primal Model BLEU scores with the Baseline BLEU scores, it lacks in 7 Indic to EN language direction i.e., Bengali, Gujarati, Hindi, Kannada, Malayalam, Odia, Punjabi-> English. Thus, these particular 7 Indic-EN cases are considered to apply different MNMT approaches to augment the scores. Various enhancement approaches are used like back-translation, domain adaptation which are discussed further in the below sub-sections. The subsequent training for these approaches is done by restoring the best checkpoint of previous model.

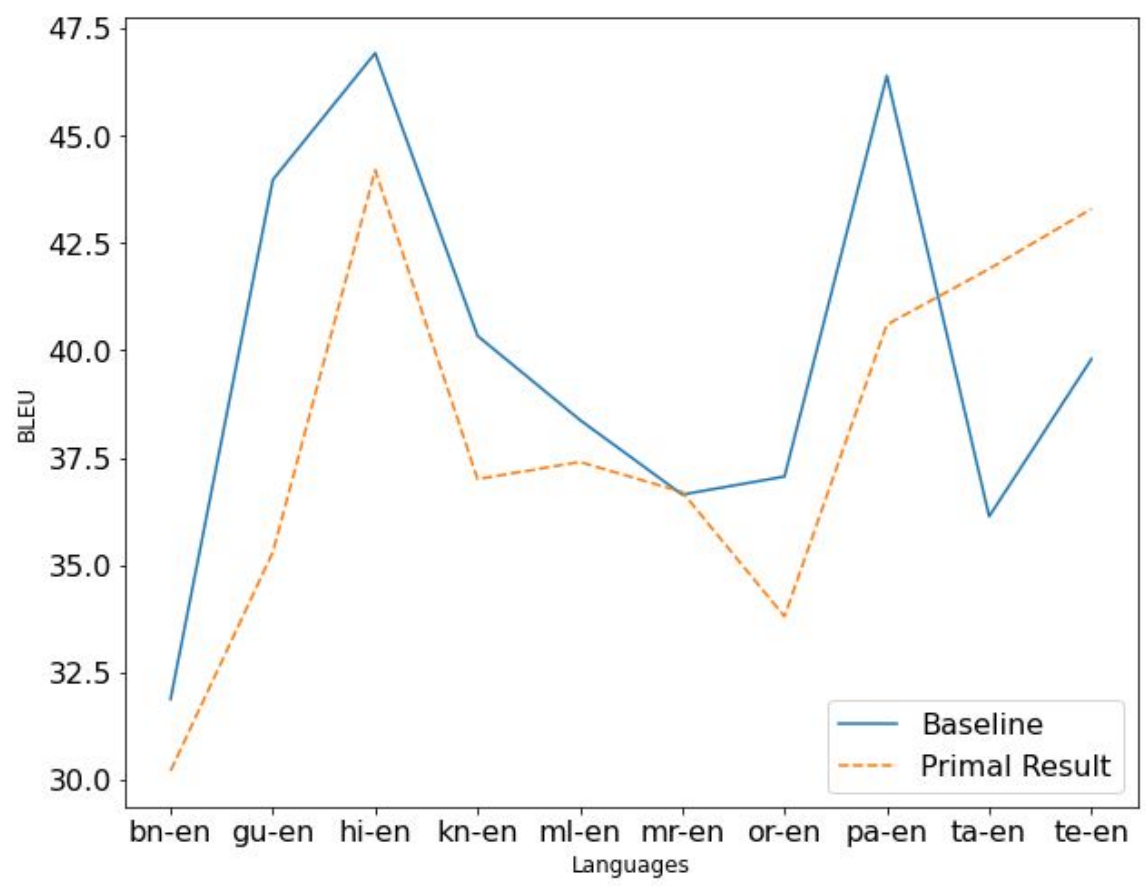

Fig. 4. Comparison of our Primal Model with Baseline

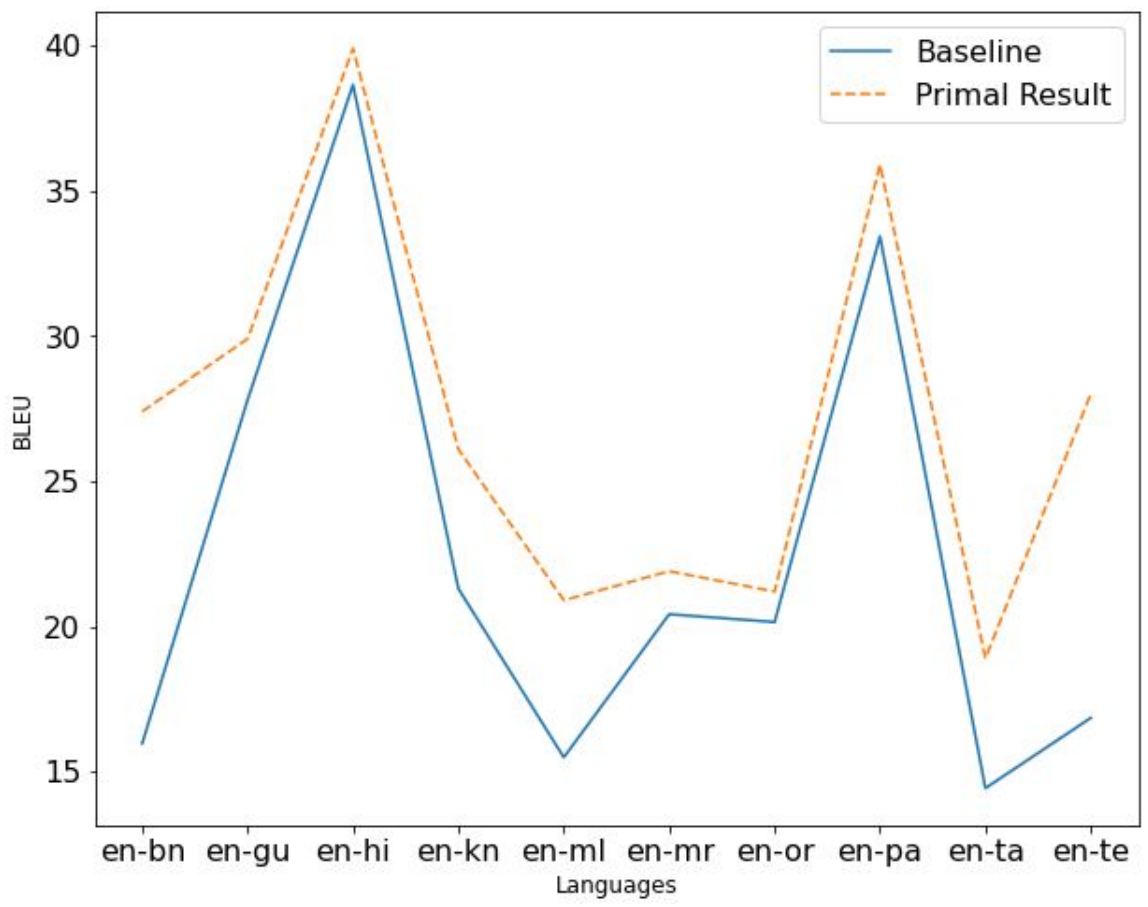

Fig. 5. Comparison of our Primal Model with Baseline

*5.2.1 Exploiting Language Relationships.* As discussed in the earlier sections, there is some similarity between all Indian languages. The majority of Indian languages are descended from the Indo-European(Aryan) and Dravidian language groups, which have been categorized as Group A and Group B respectively as referred in Table 5. Hindi, Urdu, Punjabi, Gujarati, Marathi, Oriya, Bengali and Sindhi belong to Group A whereas Telugu, Tamil, Kannada and Malayalam are Group B [57]. ILs have a rich history for its writing and scripts. These scripts are descended from the historic Brahmi script. Hindi, Bengali, Tamil, and Telugu are among the major languages that use Brahmi-scripts. However, several languages use Arabic Script. Both Arabic and Brahmi-derived scripts are used in Punjabi and Sindhi. Additionally, assuming that the populations of surrounding regions frequently mix, languages are spoken in such areas also exhibit some degree of resemblance. Hence, the effect of language mix-up technique has been considered in increasing precision by training only those languages which have some similarities among one another [38]. To apply this, methods suggested by Goyal et al[19] are followed. For example, all Indo Aryan Languages have been trained at a time to translate them to English i.e., many-to-one as well as vice versa i.e English to Indo Aryan Languages. Similarly, for languages having Devanagari Script, we train all languages in both the direction to check the effect of language relationship. One more approach is implemented by reducing the number of languages being compensated with similar languages for training the model to check the effect. For example, Hindi and Marathi is quite similar in their writing style and even Gujarati grammar also shares several similarities with Hindi. Hence, Group A languages have been trained together to check the translation quality. The same method has been applied for Kannada and Malayalam from Group B which are similar. The model for these languages has been trained to generate translations for English and vice-versa. With completion of training of both Group A and B languages, an average increment of 1.6 BLEU score has been noticed using both the techniques; classifying the languages according to their script with large number of languages as well as with most similar languages (refer to Table 4).

**Algorithm 1** : Domain Adaptation Algorithm

**Input**: Domain 1-Parallel Corpus $C_{p_1}$ Domain 2-Parallel Corpus $C_{p_2}$ Target corpus $C_t$

**Let**: No. of epoch = x
Train model $\Delta$ on $C_{p_1}$ for x epochs
Generate last checkpoint, $L_c$, for the model $\Delta$
Train model $\pi$ on $C_{p_2}$ for x epochs restoring the checkpoint $L_c$
Generate best checkpoint $L_M$ for model $\pi$
Use best checkpoint of $\pi$ to get translation results for $C_t$

**Output**: Domain adapted model $\pi$

*5.2.2 Domain Adaptation.* Domain adaptation is the process of changing a model, previously trained on any common domain, to a new domain. It is used in multi-domain problems, where systems are modified with addition of tags at the sentence/word level to provide the system with more meta-information, allowing it to produce translations using vocabulary and style that is acceptable for the domain. In order to adapt an MNMT model to a different domain, the model is typically trained on the entire parallel corpus before finetuning its parameters on a smaller in-domain corpus [43], [73]. The same model configuration as described in section 4.1 is used for implementing domain adaptation technique. However, instead of using all the 15 language pairs, 7 pairs for Indic-EN direction are considered to be used for testing this approach. A minor improvement of 0.86 average BLEU score, is achieved using this approach for Indic-EN translation direction (shown in Table 5 and Fig 6.). Algorithm 1 describes our steps for domain adaptation.

**Algorithm 2** : Back-Translation Algorithm

**Input**: Parallel Corpus $C_p$, Monolingual Corpus $C_m$, target corpus $C_t$

**Let**: $T_{r_\rightarrow} = C_p$
**loop**
  Train $Lang_1$ to $Lang_2$ model $\Delta_\rightarrow$ on $T_{r_\rightarrow}$
  With $\Delta_\rightarrow$, create new corpus $N_1$ using $C_m$
  Let $T_{r_\leftarrow} = C_p \cup N_1$
  Train $Lang_2$ to $Lang_1$ model $\Delta_\leftarrow$ on $T_{r_\leftarrow}$
  With $\Delta_\leftarrow$, create new corpus $N_2$ using $C_t$
  Let $T_{r_\rightarrow} = C_p \cup N_2$
**until** convergence condition is achieved.

**Output**: optimized & updated models $\Delta_\rightarrow$ and $\Delta_\leftarrow$

*5.2.3 Backtranslation(BT).* The statistics of massive parallel corpora, or collections of related phrases in both the source and target languages, are what machine translation relies on. Bitext has several limitations whereas there is a lot more data available in Monolingual. Here, Bitexts are a simple depiction of the source text and the translated version whereas Monolingual data refers to the data has texts for only one language. Language models have historically been trained using monolingual data, which increased the accuracy of NMT systems.

A typical method for enhancing machine translation is the BT of the monolingual corpus. BT plays a major role in more resource languages. More training data were produced using the BT technique of Sennrich et al [58] to enhance translation model performance. To produce extra synthetic parallel data from the monolingual target data, it is necessary to train a target-to-source system. This data supplements human bitext to train the intended source-to-target system. BT does not require any change of the MT training algorithms which make it simple and straightforward to use. Algorithm 2 describes our Backtranslation approach. In order to train NMT systems, we provide an approach of iterative BT, a technique for creating synthetic parallel data that is progressively improved from monolingual data. Our experiments are conducted on both low resource as well as high resource languages (according to parallel corpora as shown in Table 1) i.e., Odia , Bengali, Hindi, Kannada, Malayalam, Punjabi, Gujarati. For these languages, with the available corpora from PMI Domain [24], the monolingual corpora have been subsampled for about twice the size of the parallel training corpus. Bahdanau et al [5] neural machine translation systems are deployed to translate the monolingual data. Our configuration is similar to Sennrich et al [58]; however, the fast Marian toolkit [27] is used for training. Two different BT model approaches have been applied, such as:

- Training NMT model on parallel corpus (15k iterations stopping after 0.20 epochs (the model stops after training over 20 percent of batches with batch size of 128));
- 172K iterations where the minimal loss has been achieved i.e., convergence is achieved.

Performance of 15k iteration model is not satisfactory, and its synthetic parallel corpus does not perform better for back-translated data. Longer trained systems have significantly higher translation quality, and their synthetic parallel corpora are effective. The back-translation system that has undergone 172k iterations of training delivers noticeable advantages (+̃3.2 BLEU in Indic-EN direction for the 7 language pairs as shown in Table 5).
Hence, it is clearly visible that using all the above techniques, BLEU score for all 7 Indic to English language pair get better.

Table 5. Subsequent Stage results with evaluation metrics as BLEU score

| **Translation Direction** | **Primal Model** (BLEU Score) | **Similar Language** (BLEU Score) | **Domain adaptation** | **Back-Translation** |
|---|---|---|---|---|
| bn->en | 30.2 | 31.6 | 32.5 | 35.2 |
| gu->en | 35.3 | 37.2 | 38.5 | 43.1 |
| hi->en | 44.2 | 45.9 | 46.7 | 49.3 |
| kn->en | 37.0 | 38.6 | 39.2 | 42.3 |
| ml->en | 37.4 | 38.4 | 38.9 | 41.4 |
| or->en | 33.8 | 35.5 | 36.3 | 39.4 |
| pa->en | 40.6 | 42.3 | 43.4 | 46.8 |

## 6 FINE TUNING PRETRAINED OPUS-MT

MT model can be fine-tuned to fit a specific domain or style. A set of bilingual sentences representing the domain or style, that the MT model should adapt to, may be needed for fine-tuning. For MT fixed-domain problems, fine-tuning on robust pre-trained models (on limited, verified sets) has grown to be a preferred strategy. There are currently more than 1,000 pretrained neural MT models available in the repository of OPUS MT ranging over different language pairs. Both bilingual and multilingual models are supported by OPUS-MT [67] and its main objective is to create open-source materials and applications for MT. So, we employ **opus-mt-en-mul model** and **opus-mt-mul-en model** for fine tuning all the IL pairs in both the direction i.e. EN-XX and XX-EN. In both the terms, mul represent multiple languages, and English is represented by en. We make use of the PMI dataset [24], which is a MT indic dataset comprised of a variety of political data sources, such as news commentary and parliamentary hearings. After finetuning, it is clearly visible in Table 6 ( as well as Refer Fig. 7 and Fig. 8) that our NMT system performs better then the OPUS-MT model.

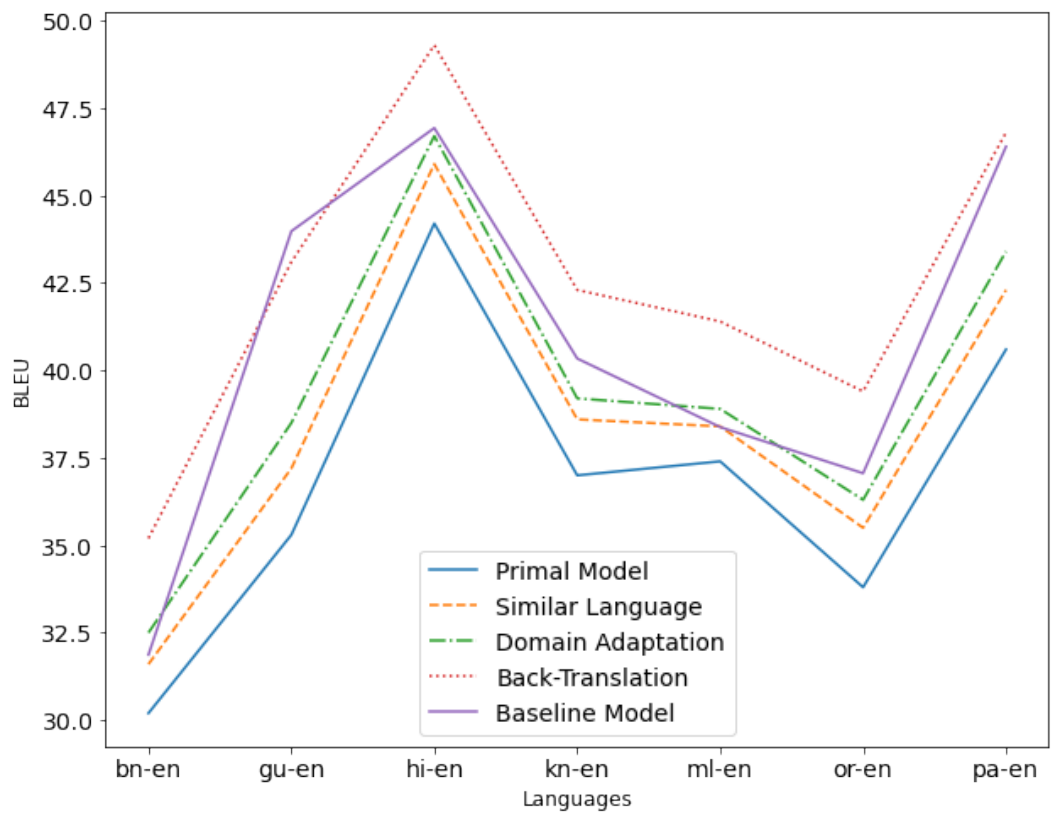

Fig. 6. Indic to English language comparison using subsequent stages(evaluation metrics as BLEU score)

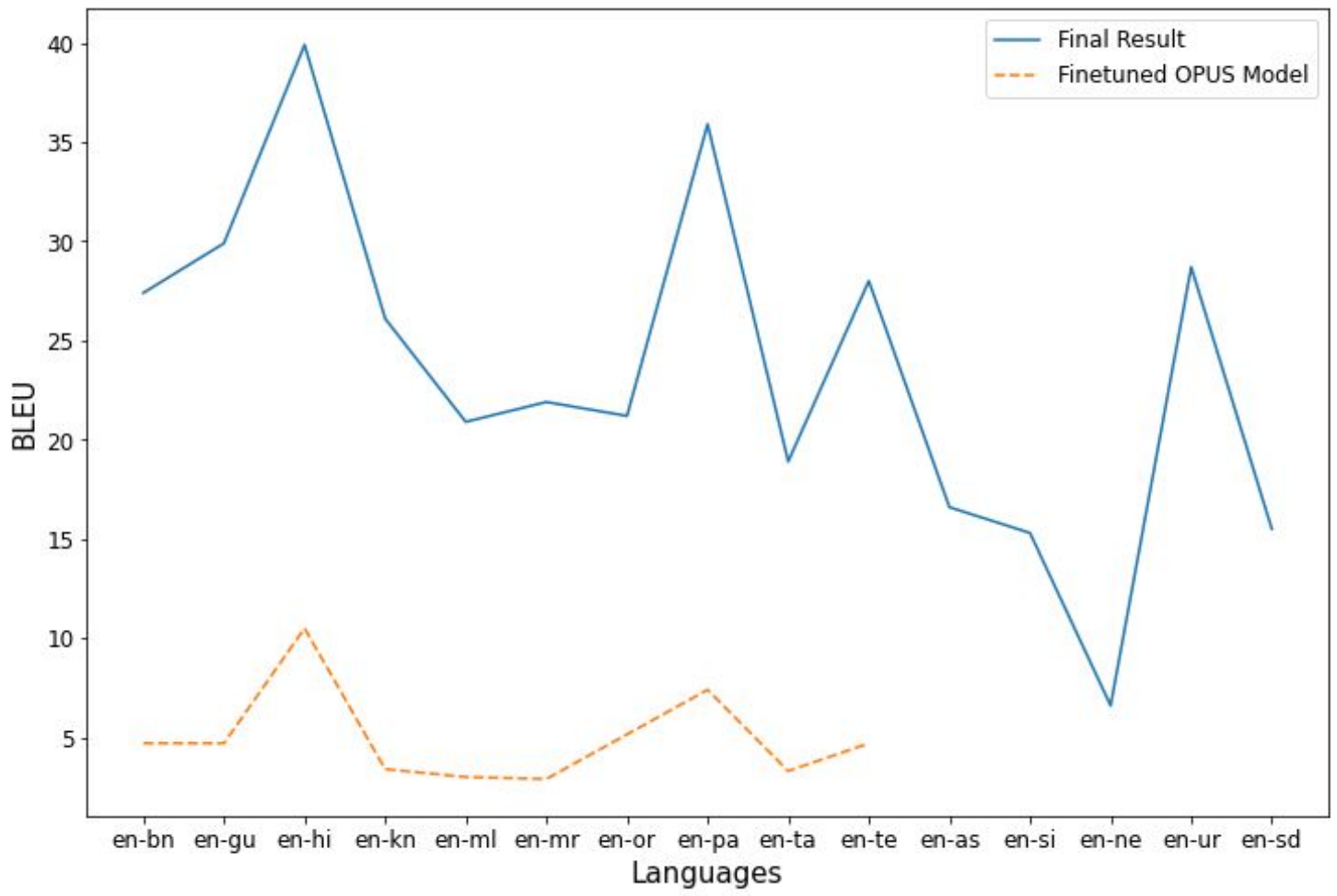

Fig. 7. Comparison of our Model with OPUS Model using evaluation metrics as BLEU score

## 7 RESULTS AND OBSERVATION

The effectiveness of the machine translation system has been evaluated using automated evaluation measures. BLEU refers to "Bilingual Evaluation Understudy" is used as an automatic evaluation metric for MT System [52]. The number of words in the MT output that match the reference translation is used to compute the BLEU. The BLEU score ranges between 0 and 1 or (0 to 100), where 0 denotes no similarities and 1 denotes identity, which is not always attainable for a model. SacreBLEU [52] is available to determine the BLEU scores of baseline models which we have used. Using WAT 2022 server [46], we have tested, and also evaluated all our translation files through BLEU score [52]. Table 7 list the BLEU score results and shows that our system is better than the baseline model [12] as well as from finetuning models for both Many to One and One to Many directions. Fig. 9 and Fig. 10 shows the comparison between our final result with baseline result. For English to IL translation, our MNMT model gives a BLEU score between 6.6 and 29.90 BLEU score, with Nepali and Gujarati receiving the lowest and the highest scores, respectively. For IL to English direction, BLEU score ranges between 14.0 to 49.3 where Nepali is having the lowest and Hindi is having the highest BLEU Score.

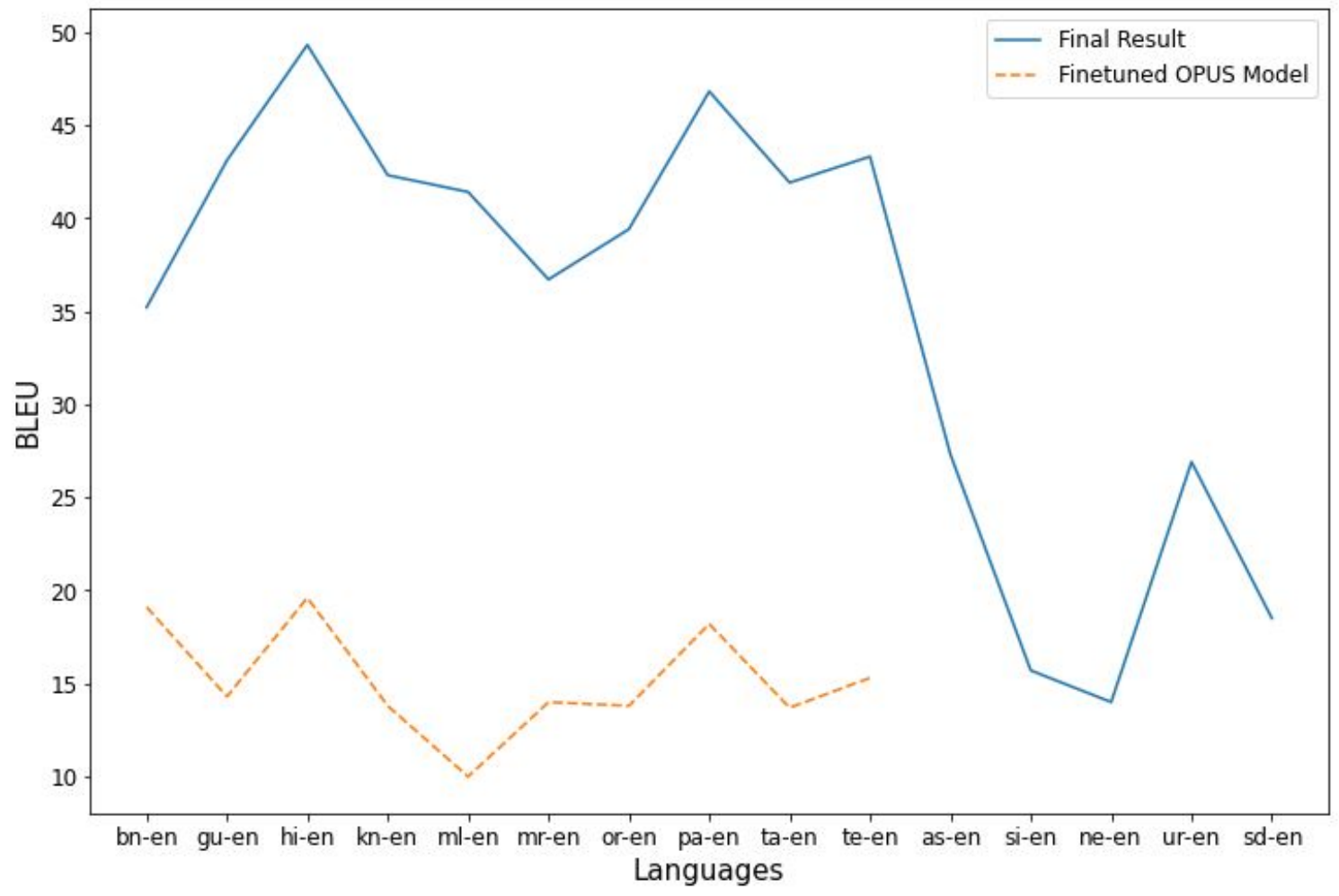

Fig. 8. Comparison of our Model with OPUS Model evaluation metrics as BLEU score

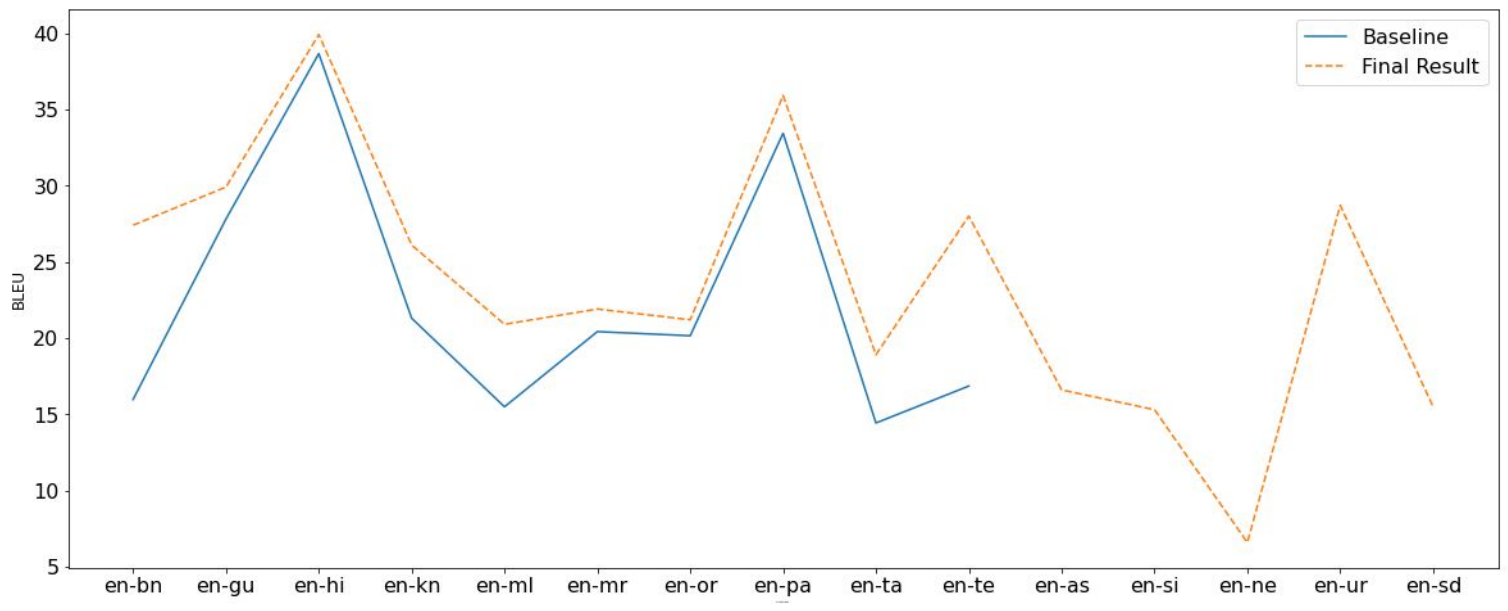

Fig. 9. Comparison of Final Result with Baseline Result evaluation metrics as BLEU score

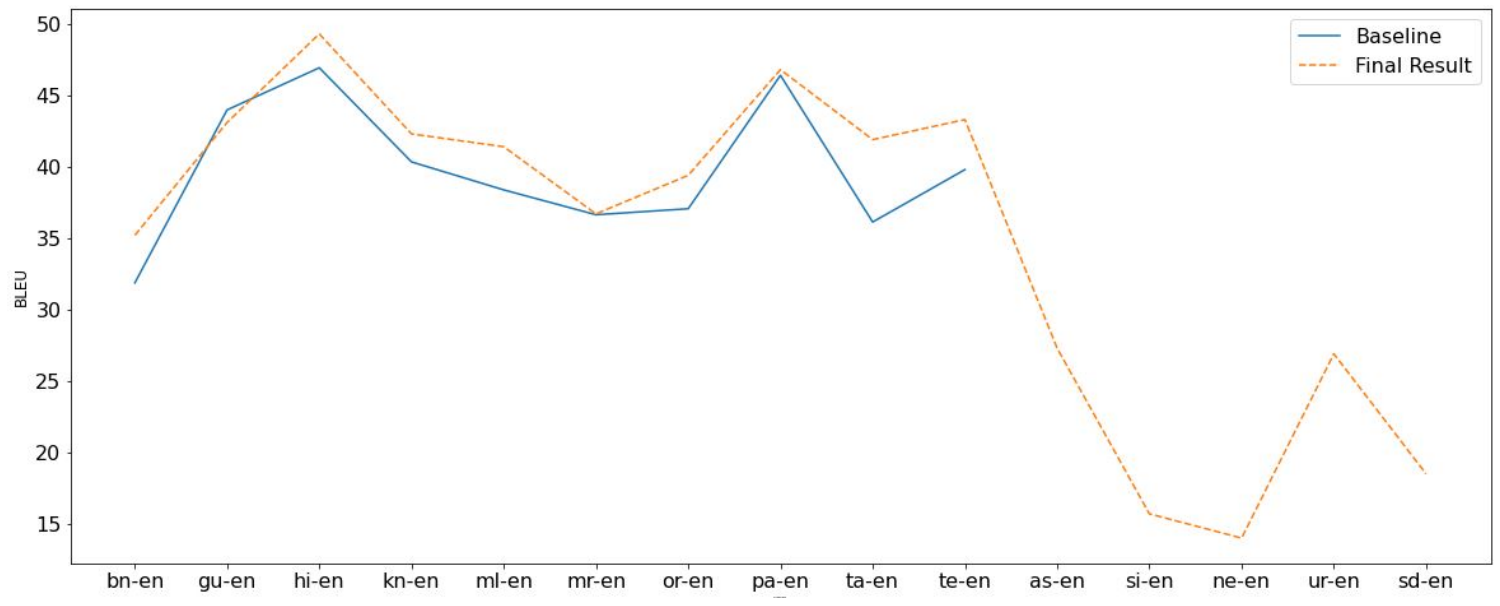

Fig. 10. Comparison of Final Result with Baseline Result with evaluation metrics as BLEU score

Table 6. Comparison with Finetuned OPUS-MT model with evaluation metrics as BLEU score

| | Final Results | Finetuned OPUS Model | | Final Results | Finetuned OPUS Model |
|---|---|---|---|---|---|
| **EN->Indic** | **BLEU** | **BLEU** | **Indic->EN** | **BLEU** | **BLEU** |
| en-bn | 27.4 | 4.7 | bn-en | 35.2 | 19.1 |
| en-gu | 29.9 | 4.7 | gu-en | 43.1 | 14.3 |
| en-hi | 39.9 | 10.5 | hi-en | 49.3 | 19.6 |
| en-kn | 26.1 | 3.4 | kn-en | 42.3 | 13.8 |
| en-ml | 20.9 | 3 | ml-en | 41.4 | 10 |
| en-mr | 21.9 | 2.9 | mr-en | 36.7 | 14 |
| en-or | 21.2 | NR* | or-en | 39.4 | 13.8 |
| en-pa | 35.9 | 7.4 | pa-en | 46.8 | 18.2 |
| en-ta | 18.9 | 3.3 | ta-en | 41.9 | 13.7 |
| en-te | 28 | 4.7 | te-en | 43.3 | 15.3 |
| en-as | 16.6 | NR* | as-en | 27.3 | NR* |
| en-si | 15.3 | 7.4 | si-en | 15.7 | NR* |
| en-ne | 6.6 | NR* | ne-en | 14 | NR* |
| en-ur | 28.7 | NR* | ur-en | 26.9 | NR* |
| en-sd | 15.5 | NR* | sd-en | 18.5 | NR* |

NR* - NOT recorded. The OPUS-MT executable model not found for these languages.

Table 7. Comparison of Final Result with Baseline Result with evaluation metrics as BLEU score

| | Baseline | Final Result | | Baseline | Final Result |
|---|---|---|---|---|---|
| **EN->Indic** | **BLEU** | **BLEU** | **Indic->EN** | **BLEU** | **BLEU** |
| en-bn | 15.97 | **27.40** | bn-en | 31.87 | **35.2** |
| en-gu | 27.80 | **29.90** | gu-en | 43.98 | **43.1** |
| en-hi | 38.65 | **39.9** | hi-en | 46.93 | **49.3** |
| en-kn | 21.30 | **26.10** | kn-en | 40.34 | **42.3** |
| en-ml | 15.49 | **20.90** | ml-en | 38.38 | **41.4** |
| en-mr | 20.42 | **21.90** | mr-en | 36.64 | **36.7** |
| en-or | 20.15 | **21.20** | or-en | 37.06 | **39.4** |
| en-pa | 33.43 | **35.90** | pa-en | 46.40 | **46.8** |
| en-ta | 14.43 | **18.90** | ta-en | 36.13 | **41.9** |
| en-te | 16.85 | **28.0** | te-en | 39.8 | **43.3** |
| en-as | - | **16.6** | as-en | - | **27.3** |
| en-si | - | **15.3** | si-en | - | **15.7** |
| en-ne | - | **6.6** | ne-en | - | **14** |
| en-ur | - | **28.70** | ur-en | - | **26.9** |
| en-sd | - | **15.50** | sd-en | - | **18.5** |

## 7.1 Translation Files

This subsection give an overview about the translation of all the languages generated by our model.

(1) **English to Assamese**

**English:** Until 1960, Brzezinski worked as an advisor to John F. Kennedy, then to Lyndon B. Kennedy. on behalf of the Johnson administration.
**Generated:** ১৯৬০ চনত ব্ৰজেজিনস্কি জন এফ কেনেডিৰ উপদেষ্টা আৰু তাৰ পিছত লিন্ডন বি . জনছনৰ প্ৰশাসনে কাম কৰিছিল ।
**Reference:** ১৯৬০ চনলৈকে, ব্ৰজেন্সিকিয়ে উপদেষ্টাৰূপে জন এফ কেনেডিৰ হৈ কাম কৰিছিল তাৰপিছত তেওঁ লিনডন বি. জনচন প্ৰশাসনৰ হৈ কাম কৰিছিল।

(2) **English to Bengali**

**English:** Until 1960, Brzezinski worked as an advisor to John F. Kennedy, then to Lyndon B. Kennedy. on behalf of the Johnson administration.
**Generated:** ১৯৬০ - এর দশকে , ব্রিজেজিনস্কি জন এফ কেনেডির উপদেষ্টা এবং তারপরে লিন্ডন বি . জনসন প্রশাসন হিসাবে কাজ করেছিল ।
**Reference:** 1960 সাল জুড়ে, ব্রজেজিনস্কি জন এফ. কেনেডির জন্য উপদেষ্টা হিসেবে কাজ করেছিলেন এবং পরবর্তীতে লিন্ডন বি. জনসন প্রশাশন এর সথে।।

(3) **English to Gujarati**

**English:** The discovery also provides a deeper understanding of the evolution of bird feathers.
**Generated:** આ શોધ પક્ષીઓમાં પાંખો ઉત્ક્રાંતિની પણ સમજણ આપે છે .
**Reference:** આ શોધ પક્ષીઓના પીછાની ઉત્ક્રાંતિ ની ગૂઢ સમજણ પણ આપે છે.

(4) **English to Hindi**

**English:** In late 2017, Siminoff appeared on the shopping television channel QVC.
**Generated:** 2017 के अंतें , सिमिनॉफ शॉपिंग टेलीविजन चैनल क्यूवीसी पर दिखाई दिया ।
**Reference:** 2017 के अंत में, सिमिनॉफ़ शॉपिंग टेलीविज़न चैनल QVC में दिखाई दिए.

(5) **English to Kannada**

**English:** The discovery also offers insight into the evolution of bird feathers.
**Generated:** ಈ ಸಂಶೋಧನೆಯು ಪಕ್ಷಿಗಳಲ್ಲಿನ ರೆಕ್ಕೆಗಳ ಬೆಳವಣಿಗೆಯ ಬಗ್ಗೆಯೂ ಒಳನೋಟವನ್ನು ನೀಡುತ್ತದೆ .
**Reference:** ಆವಿಷ್ಕಾರವು ಪಕ್ಷಿ ಗರಿಗಳ ವಿಕಾಸದ ಒಳನೋಟವನ್ನು ಸಹ ನೀಡುತ್ತದೆ.

(6) **English to Malayalam**

**English:** In late 2017, Siminoff appeared on shopping television channel QVC.
**Generated:** 2017 അവസാനം , സൈമിനോഫ് ഷോപ്പിംഗ് ടെലിവിഷൻ ചാനലായ QVCൽ പ്രത്യക്ഷപ്പെട്ടു .
**Reference:** 2017-ന്റെ അവസാനത്തിൽ, സിമിനോഫ് ഷോപ്പിംഗ് ടെലിവിഷൻ ചാനലായ

ക്യുവിസിയിൽ പ്രത്യക്ഷപ്പെട്ടു.

(7) **English to Marathi**

**English:** Ring settled a lawsuit with a rival security company, ADT Corporation.
**Generated:** रिंगने स्पर्धात्मक सुरक्षा कंपनी , एडीटी कॉर्पोरेशनसोबत खटला देखील सुलभ केला .
**Reference:** रिंगने प्रतिस्पर्धी सुरक्षा कंपनी, एडीटी कॉर्पोरेशनसोबत खटल्याचा निकाल लावला.

(8) **English to Oriya**

**English:** Gosling and Stone received nominations for Best Actor and Best Actress, respectively.
**Generated:** ଗୋସ୍ଲିଙ୍ଗ ଏବଂ ଷ୍ଟୋନ ଯଥାକ୍ରମେ ଶ୍ରେଷ୍ଠ ଅଭିନେତା ଏବଂ ଅଭିନେତ୍ରୀ ଭାବେ ନାମାଙ୍କନ ପାଇଥିଲେ ।
**Reference:** ଗୋସଲିଂ ଏବଂ ଷ୍ଟୋନ୍ ଯଥାକ୍ରମେ ଶ୍ରେଷ୍ଠ ଅଭିନେତା ଓ ଶ୍ରେଷ୍ଠ ଅଭିନେତ୍ରୀ ପାଇଁ ନାମାଙ୍କନ ଗ୍ରହଣ କରିଥିଲେ ।

(9) **English to Punjabi**

**English:** Now we have 4-month-old mice that are non-diabetic that are used for diabetics", he added.
**Generated:** " ਹੁਣ ਸਾਡੇ ਕੋਲ 4 ਮਹੀਨੇ ਦੇ ਚੂਹੇ ਹਨ ਜੋ ਗੈਰ - ਡਾਇਬਟੀਜ਼ ਹਨ ਜੋ ਪਹਿਲਾਂ ਡਾਇਬਟੀਜ਼ ਹੁੰਦੇ ਸਨ , " ਉਸਨੇ ਕਿਹਾ
**Reference:** ਹੁਣ ਸਾਡੇ ਕੋਲ 4 ਮਹੀਨਿਆਂ ਦਾ ਚੂਹੇ ਹਨ ਜੋ ਗੈਰ-ਸੁਗਰ ਰੋਗੀ ਹਨ ਜੋ ਸੁਗਰ ਰੋਗੀਆਂ ਲਈ ਵਰਤੇ ਜਾਂਦੇ ਹਨ", ਉਸਨੇ ਜੋੜਿਆ।

(10) **English to Tamil**

**English:** The Iraq Study Group delivered its report at 12.00 GMT today.
**Generated:** : ஈராக் ஆய்வு குழு தனது அறிக்கையை இன்று 12.00 GMT க்கு சமர்ப்பித்தது
**Reference:** ஈராக் ஆய்வுக் குழு, தனது அறிக்கையை இன்று 12.00 GMT- க்கு வழங்கியது.

(11) **English to Telugu**

**English:** He built a WiFi doorbell. He said.
**Generated:** అతను ఒక వైఫై డోర్ బెల్ నిర్మించాడు .
**Reference:** అతను WiFi డోర్ బెల్ నిర్మించాడు. అని చెప్పాడు.

(12) **English to Sinhala**

**English:** The crew of the ship was divided into three main sections.
**Generated:** නැවේ කාර්ය මණ්ඩලය ප්රධාන කොටස් තුනකට වර්ගීකරණය කර ඇත .
**Reference:** නැවෙහි කාර්යය මණ්ඩලය ප් රධාන කොටස් තුනකට වනේ කොට තිබුණි.

(13) **English to Sindhi**

**English:** In late 2017, Summons of Shopping appeared on the television channel QVC.
**Generated:** 2017 ع جي آخر ۾، سمنس فونيم ۾ ٽي وي جي ليوڊ لاڊ QVC ۾ ظاهر ٿيو.

**Reference:** 2017 جي آخر م، سمن آف شاپنگ ڈيليوزن چينل QVC تي ظاهر ڈي.

(14) **English to Urdu**

**English:** : He said he had invented a Wi-Fi doorbell.
**Generated:** س نےدروازے کی گھنٹی بجائی ، اس نےکہا
**Reference:** س نےکہا کہ اس نےایک وائی فائی ڈور بیل اختراع کیا ہے

(15) **English to Nepali**

**English:** He said he had invented a Wi-Fi doorbell.
**Generated:** उनले वाईफाई डोरबेल आविष्कार गरेको बताए
**Reference:** उनले वाईफाई डोरबेल आविष्कार गरेको बताए

(1) **Assamese to English**

**Assamese:** এই যাত্ৰাত বিভিন্ন সময়ত ইৱাছাকি বিপদত পৰিছিল।
**Generated:** On this trip, Iwasaki was in danger at various times.
**Reference:** During his trip, Iwasaki ran into trouble on many occasions

(2) **Bengali to English**

**Bengali:**অনুসন্ধানটি পাখির পালকের বিবর্তনের বিষয়েও পরিজ্ঞান দেয়
**Generated:** the discovery also gives insight into the evolution of bird feathers
**Reference:** The find also grants insight into the evolution of feathers in birds.

(3) **Gujarati to English**

**Gujarati:** તેમણે વાઈફાઈ ડોર બેલ બનાવ્યો હતો, એમ તેમણે કહ્યું હતું.
**Generated:** he made a wifi bell , he said .
**Reference:** He built a WiFi door bell, he said.

(4) **Hindi to English**

**Hindi:** रिंग ने अपनी प्रतिस्पर्धी कंपनी ADT कॉर्पोरेशन के साथ एक मुकदमा निपटाया है.
**Generated:** ring has settled a lawsuit with its rival adt corporation .
**Reference:** Ring also settled a lawsuit with competing security company, the ADT Corporation

(5) **Kannada to English**

**Kannada:** ಈ ಮೊದಲು ಡಯಾಬೆಟಿಕ್ ಆಗಿದ್ದ ಆದರೆ ಈಗ ಡಯಾಬೆಟಿಕ್ ಅಲ್ಲದ 4 ತಿಂಗಳ ಇಲಿಗಳು ನಮ್ಮ ಬಳಿ ಇವೆ," ಎಂದು ಅವನು ಹೇಳಿದ.
**Generated:** Earlier it was diabetic , but now we have 4 - month mice without diabetes , he said .

**Reference:** We now have 4-month-old mice that are non-diabetic that used to be diabetic," he added.

(6) **Malayalam to English**

**Malayalam:** അദ്ദേഹം ഒരു WiFi ഡോർ ബെൽ ഉണ്ടാക്കിയെന്ന് അവൻ പറഞ്ഞു.
**Generated:** he said he made a wifi door bell.
**Reference:** He built a WiFi door bell, he said.

(7) **Marathi to English**

**Marathi:** रविवारी उशिरा संयुक्त राष्ट्रसंघाचे अध्यक्ष डोनाल्ड ट्रम्प यांनी, प्रेस सुरक्षेच्या मार्फत दिलेल्या विधानात यूएस सैनिक सीरिया सोडत असल्याची घोषणा केली.
**Generated:** US president donald trump announced that his troops were leaving syria .
**Reference:** Late on Sunday, the United States President Donald Trump, in a statement delivered via the press secretary, announced US troops would be leaving Syria.

(8) **Oriya to English**

**Oriya:** ଗୋସଲିଂ ଏବଂ ଷ୍ଟୋନ୍ ଯଥାକ୍ରମେ ଶ୍ରେଷ୍ଠ ଅଭିନେତା ଓ ଶ୍ରେଷ୍ଠ ଅଭିନେତ୍ରୀ ପାଇଁ ନାମାଙ୍କନ ଗ୍ରହଣ କରିଥିଲେ ।
**Generated:** Gosling and stone were nominated for best actor and best actress respectively.
**Reference:** Gosling and Stone received nominations for Best Actor and Actress respectively.

(9) **Punjabi to English**

**Punjabi:** ਆਪਣੀ ਯਾਤਰਾ ਦੇ ਦੌਰਾਨ, ਈਵਾਸਾਕੀ ਕਈ ਵਾਰ ਮੁਸੀਬਤ ਵਿੱਚ ਫਸੇ।
**Generated:** during his journey , iwasaki was sometimes in trouble .
**Reference:** During his trip, Iwasaki ran into trouble on many occasions.

(10) **Tamil to English**

**Punjabi:** ஈராக் ஆய்வுக் குழு, தனது அறிக்கையை இன்று 12.00 GMTக்கு வழங்கியது.
**Generated:** the iraqi research team today submitted its report to the 12 : 00 gmt .
**Reference:** The Iraq Study Group presented its report at 12.00 GMT today.

(11) **Telugu to English**

**Telugu :** ఆదివారం అర్ధరాత్రి అమెరికా అధ్యక్షుడు డొనాల్డ్ ట్రంప్ పత్రికా కార్యదర్శి ద్వారా విడుదల చేసిన ఒక ప్రకటనలో అమెరికా దళాలు సిరియా నుంచి వైదొలుగుతున్నట్లు ప్రకటించారు.
**Generated:** US president donald trump announced in a statement that his troops were leaving syria .
**Reference:** Late on Sunday, the United States President Donald Trump, in a statement delivered via the press secretary, announced US troops would be leaving Syria.

(12) **Sinhala to English**

**Sinhala :** නැව මත සිටි සංඛ් යාවනේ තුනනේ එකකට අඩු ප් රමාණයක් පමණක් දිවිගලවා ගන තිබුණි.
**Generated:** Only one third of the population on ship had survived.
**Reference:** From the amount onboard the ship about one third had saved their lives.

(13) **Sindhi to English**

**Sindhi:** گورنر جي آفيس ٻڌايو ته زخمي ٿيل ماڻهن ۾ اڻويهه پوليس وارا هئا.
**Generated:** According to the governor's office, there were 19 police officers in the affected people.
**Reference:** The governor's office said nineteen of the injured were police officers.

(14) **Urdu to English**

**Urdu:** ریان گوسلنگ اور ایما سٹون کی اداکاری والی فلم کو تمام بڑے زمروں میں نامزدگی موصول ہوئی
**Generated:** Ryan Gosling and Emma stone's acting films get nominated in all major categories.
**Reference:** The movie, featuring Ryan Gosling and Emma Stone, received nominations in all major categories.

(15) **Nepali to English**

**Nepali:** उनले वाईफाई डोरबेल आविष्कार गरेको बताए
**Generated:** He said he had invented Wi-Fi doorbell.
**Reference:** He said he had invented a Wi-Fi doorbell.

## 8 CONCLUSION

We present a multilingual neural machine translation system for 15 ILs i.e., English to Indic and vice versa which offers better results in comparison to existing MT systems for IL. Different techniques are explored in this paper such as finetuning, exploiting Language relationships, back translation and domain adaptation. Experiments on Samantar datasets reveal that our methods significantly outperform the best individual systems and cutting-edge traditional system combination techniques. Results, in comparison with the pretrained OPUS-MT NMT model for the majority of language pairs using the PMI corpus, confirm that our MNMT model performed better than the pretrained model. The effectiveness of language relatedness is explored, analysed and validated to be useful for the low resource language(s), with support of high resource language(s), for ascertaining a qualitative translation by increasing the corpus. Further, our results show that Back Translation(BT) enhances the translation quality of both sources and targets over original sentences (by +3.2 BLEU in both directions) being augmented with domain adaptation. However, even with the use of a reasonable number of data filtering methods, a thorough inspection on some sentences (from the filtered dataset for all the languages) makes it evident that noise still contaminates the training data. Such trend demands for futuristic research in effective data filtering techniques towards betterment of MT performance. Even though MT has made some significant development but still there is ample scope for improvements. This paper suggests various methods of MT to ensure better performance than the standard baseline model. And, for low-resource languages like ILs, substantial progress is needed in terms of quality of translation. In order to take it to the next level, futuristic works may focus on experimenting on the combination of approaches with a proper trade-off among various individualistic methods/systems.